\documentclass[lettersize,journal]{IEEEtran}
\usepackage{amsmath,amsfonts}
\usepackage{algorithmic}
\usepackage{algorithm}
\usepackage{array}
\usepackage{amsmath}
\usepackage{amssymb}
\usepackage{caption}
\usepackage{subcaption}
\usepackage{textcomp}
\usepackage{stfloats}
\usepackage{url}
\usepackage{verbatim}
\usepackage{graphicx}
\usepackage{cite}
\usepackage{multicol}
\usepackage{color}
\begin{document}

\title{Dual-Segment Clustering Strategy for Hierarchical Federated Learning in Heterogeneous Wireless Environments}

\author{Pengcheng Sun,
Erwu Liu,
Wei Ni, \IEEEmembership{Fellow,~IEEE,}
Kanglei Yu,
Xinyu Qu,
Rui Wang,
Yanlong Bi,
Chuanchun Zhang,
and Abbas Jamalipour, \IEEEmembership{Fellow,~IEEE}
        % <-this % stops a space
\thanks{
% This work is supported in part by grants from the National Natural Science Foundation of China (No. 42171404, No.42225401) and Shanghai Engineering Research Center for Blockchain Applications And Services (No. 19DZ2255100).
This work is supported in part by grants from the National Natural Science Foundation of China (No. 42171404, No. 82070920), Shanghai Engineering Research Center for Blockchain Applications And Services (No. 19DZ2255100), and Key Disciplines of the Sixth Cycle of Tongji Hospital Affiliated to Tongji University (ZDPY24-YK).
}
\thanks{P. Sun, E. Liu, K. Yu, X. Qu and R. Wang are with the College of Electronics and Information Engineering, Tongji University. E. Liu and Y. Bi are with the Department of Ophthalmology, Tongji Hospital, Tongji University. E-mails: pc\_sun2020@tongji.edu.cn, erwu.liu@ieee.org, 2152206@tongji.edu.cn, xinyuqu@tongji.edu.cn, ruiwang@tongji.edu.cn, biyanlong@tongji.edu.cn.
}% <-this % stops a space
\thanks{W. Ni is with Data61, CSIRO, Australia. E-mail: wei.ni@ieee.org.
}
\thanks{C. Zhang is with Guangzhou Huatu Information Technology Co., Ltd. E-mail: zhangcc@huatugz.com.
}
\thanks{A. Jamalipour is with the School of Electrical and Computer Engineering, The University of Sydney, Australia, E-mail: a.jamalipour@ieee.org.
}
\thanks{Corresponding author: Erwu Liu.}
}

% \markboth{Submitted to IEEE Wireless Communicaitons Letters on \today}%


\maketitle

\begin{abstract}
Non-independent and identically distributed (Non-IID) data adversely affects federated learning (FL) while heterogeneity in communication quality can undermine the reliability of model parameter transmission, potentially degrading  wireless FL convergence.
This paper proposes a novel dual-segment clustering (DSC) strategy that jointly addresses communication and data heterogeneity in FL. This is achieved by defining a new signal-to-noise ratio (SNR) matrix and information quantity matrix to capture the communication and data heterogeneity, respectively. The celebrated affinity propagation algorithm is leveraged to iteratively refine the clustering of clients based on the newly defined matrices effectively enhancing model aggregation in heterogeneous environments.
The convergence analysis and experimental results show that the DSC strategy can improve the convergence rate of wireless FL and demonstrate superior accuracy in heterogeneous environments compared to classical clustering methods.
\end{abstract}


\section{Introduction}
\IEEEPARstart{F}{ederated} learning (FL) shares the model parameters or gradients instead of the raw data to effectively reduce communication load while preserving data privacy\cite{FL1,FedAvg}. Heterogeneous environments, including non-independent and identically distributed (Non-IID) data and heterogeneous communication quality, can substantially compromise the performance of FL aggregation\cite{noniid2,noniidcomm}.

Clustering clients before aggregation is an effective way to improve the aggregation efficiency and accuracy of FL. 
% Duan \textit{et al.}\cite{hierarchical} proposed a hierarchical FL framework by setting up a proxy server within a group to aggregate the parameters of the members and then uploading them to the parameter server to update the global parameters. 
Duan \textit{et al.}\cite{hierarchical} proposed a hierarchical FL framework where a proxy server aggregates clients' parameters within a group before uploading them to the parameter server for global updates.
% In the hierarchical framework proposed in\cite{dataquantization}, data quantization and sequence learning were used within groups to improve the aggregation efficiency of FL under a Non-IID data setting. 
In \cite{dataquantization}, data quantization and sequence learning were used within groups to improve the aggregation efficiency of FL under a Non-IID data setting. 
% Most existing grouping methods only alleviate the impact of data heterogeneity on FL aggregation efficiency. However, the heterogeneity of communication affects the clients' transmission quality and, hence, the aggregation performance of FL.
However, most existing grouping methods only address data heterogeneity, overlooking communication heterogeneity, which affects transmission quality and ultimately impacts  wireless FL aggregation performance.

Works in \cite{comgroup1,comgroup2} designed clustering strategies based on communication cost, and did not consider the data heterogeneity of each group. The authors of \cite{comdata1,comdata2,comdata3} comprehensively captured the clients' communication capability and the heterogeneity of data while clustering. They reduced transmission delay, instead of addressing the impact of the communication quality on FL aggregation.

% This paper proposes a new dual-segment clustering (DSC) strategy to address both the Non-IID characteristics of local data and the heterogeneity of communication quality in FL. First, the communication conditions of clients, such as signal-to-noise ratios (SNRs), can differ substantially, leading to inconsistent communication quality in model updates. To mitigate this, an SNR matrix is constructed to capture the communication quality of each client, ensure that clients with similar communication capabilities are grouped together, thereby improving the efficiency of information transmission. Second, within each primary cluster, the heterogeneity of local data distributions further complicates the aggregation process, as Non-IID data degrades the global model's performance.  To handle this, a new information quantity matrix of local data distribution is built to create secondary clusters that address this variance in data.  Finally, for scattered clients that remain ungrouped, the principle of Euclidean distance-based proximity is adopted to integrate them into their respective \textit{nearby groups}, ensuring that all clients are accounted for in the learning process. 
% By implementing the DSC strategy, the data distribution between the groups exhibits enhanced similarity.  Moreover, the communication conditions among the clients within each group remain relatively consistent, significantly enhancing FL aggregation efficiency and accuracy. 
This paper proposes a new dual-segment clustering (DSC) strategy, which addresses the heterogeneity in both data and communication capability of wireless FL.  The innovation lies in the joint consideration of these two aspects to enhance client clustering. 
Specifically, we interpret this clustering problem as a multi-dimensional balancing problem. We define a signal-to-noise ratio (SNR) matrix to quantize the impart of communication quality and an information quantity matrix to measure the local data distribution heterogeneity. By using the affinity propagation algorithm\cite{affinity} designed to solve complex adaptive clustering problem, we interactively refine cluster assignments based on the two matrices until convergence. With the excellent effectiveness of the affinity propagation algorithm, our approach can balance data heterogeneity and communication during clustering.
% i.e., our approach integrates both dimensions by employing a two-stage clustering process.  Specifically, we alternate between clustering based on communication quality, represented by a proposed signal-to-noise ratio (SNR) matrix, and local data distribution heterogeneity, captured via a proposed information quantity matrix.  This dual-segment framework is implemented using the affinity propagation algorithm\cite{affinity}, ensuring optimal clustering through iterative refinement.  By first forming clusters that account for the clients' communication conditions—grouping those with similar SNRs—we improve communication efficiency.  Subsequently, we address the challenges posed by Non-IID data within these clusters by further subdividing clients based on the difference in their local data distributions.  Finally, clients not fitting into any cluster are integrated based on Euclidean distance proximity, ensuring full participation in the learning process.
This method effectively manages the trade-off between communication and data heterogeneity, improving model aggregation and offering a meaningful advancement in the design of heterogeneous wireless FL.

The convergence upper bound of wireless FL under the new DSC strategy is analyzed, showing that this strategy reduces the noise and bias in gradient updates under heterogeneous conditions. Experimental results show that the proposed DSC algorithm achieves 20.28\% and 21.42\% accuracy improvement on the MNIST and Fashion-MNIST datasets, respectively.
To our knowledge, this is the first clustering strategy addressing both data and communication heterogeneity in wireless FL.

The remainder of this paper is structured in the following manner: Section II illustrates the system model, including the group-based hierarchical FL aggregation and wireless communication channel. Section III elaborates on the proposed DSC strategy and analyzes its convergence. The simulation results are presented in Section~IV. Section V draws the conclusions.

\section{System Model}

We consider an FL system consisting of an $N_a$-antenna BS (serving as the parameter server) and \textit{K} single-antenna clients. The $k$-th client ($k=1,\cdots,K$) has its local data set $\mathcal{D}_k$. Consider an FL algorithm with the input data vector $\boldsymbol{x}_{ks}\in\mathbb{R}^d$ and the output $y_{ks}\in\mathbb{R}$, where $s\in \{1,\cdots,|\mathcal{D}_k|\}$ is the index of a data sample and $|\cdot|$ stands for cardinality. Let $\boldsymbol{w}_k$ be the model parameters of the local model trained at the $k$-th client. 

\subsection{Learning Model}

% For the clients, the goal of local training is to find the optimal learning model $\boldsymbol{w}^\ast$ that minimizes the local training loss. 

% For brevity, we rewrite $f_k\left(\boldsymbol{x}_{ks},y_{ks};\boldsymbol{w}\right)$ as $f_k\left(\boldsymbol{w}\right)$. 

To achieve the minimum global loss function, FL conducts multiple rounds of gradient transmission until convergence. The local gradient of the model $\boldsymbol{w}\in\mathbb{R}^q$ (with the model size $q$) on $\mathcal{D}_k$ in the $t$-th communication round is given by
\begin{equation}
\nabla F_k\left(\boldsymbol{w}^{[t]}\right)=\frac{1}{|\mathcal{D}_k|}\sum_{\left(\boldsymbol{x}_{ks},y_{ks}\right)\in\mathcal{D}_k} \nabla f_k\left(\boldsymbol{x}_{ks},y_{ks};\boldsymbol{w}^{[t]}\right)\label{gradient}
,\end{equation}
where $f_k\left(\boldsymbol{x}_{ks},y_{ks};\boldsymbol{w}\right)$ is the sample loss per the $s$-th sample.

Clustering is performed to group the clients into $L$ groups. At each communication round, the gradients from the clients in each group are first synchronously aggregated at the nominated leader of the group, and then the BS aggregates the gradients from all group leaders, as given by
\begin{equation}
    \nabla F\left(\boldsymbol{w}^{[t]}\right)=\sum_{l=1}^{L}{G_l  [\sum_{k=1}^{K_l}G_k  \nabla F_k\left(\boldsymbol{w}^{[t]}\right)]},
\label{aggregationw}
\end{equation}
where $l= 1,\cdots,L$ is the index of a group, $G_k$ is the intra-group aggregation coefficient of client $k$,  $G_l$ is the inter-group aggregation coefficient of group leader~$l$, and $K_l$ is the number of clients within the $l$-th group satisfying $K=\sum_{l=1}^{L}{K_l}$. 

Each group includes as many sample labels as possible. Hence, an aggregation coefficient dedicated to the Non-IID case is used within the group due to significant differences in the data distribution, as given by \cite{aggragatewei}
\begin{equation}
    G_k = \frac{|\mathcal{D}_k| e^{f(\theta_k^{[t]})}}{\sum_{i=1}^{K_l}{|\mathcal{D}_i| e^{f(\theta_i^{[t]})}}},
\end{equation}
where $f(\theta_i^{[t]})=1-e^{-e^{-(\theta_i^{[t]}-1)}}$, $\theta_i^{[t]}=\text{arccos}\frac{\langle \nabla F_l\left(\boldsymbol{w}^{[t]}\right), \nabla F_k\left(\boldsymbol{w}^{[t]}\right)\rangle}{\|\nabla F_l\left(\boldsymbol{w}^{[t]}\right)\|\cdot \|\nabla F_k\left(\boldsymbol{w}^{[t]}\right)\|}$, and $G_l = |\mathcal{D}_l|/\sum_{l=1}^{L}{|\mathcal{D}_l|}$ is used for small difference among the groups.

Finally, the global model at the BS is updated by 
\begin{equation}
    \boldsymbol{w}^{[t+1]}= \boldsymbol{w}^{[t]} - \lambda \cdot \nabla F\left(\boldsymbol{w}^{[t]}\right),
\label{global}
\end{equation}
where $\lambda$ is the learning rate.

\subsection{Communication Model}

% Non-Orthogonal Multiple Access (NOMA) \cite{NOMA} is considered in this paper to support multiple users and the BS to transmit data simultaneously through superimposed wireless channels, effectively alleviating communication network congestion caused by multiple users in FL \cite{NOMA1}. No inter-client interference is considered in this paper.

% Consider a block fading channel, where the channel coefficients remain invariant during the whole FL training process.
Let $\boldsymbol{h}_{k}\in\mathbb{C}^{N_a\times1}$ denote the channel coefficient vector of the direct channel from the $k$-th client to the BS, and $h_{jk}\in\mathbb{C}$ be the channel coefficient from the $k$-the client to the $j$-the client. In the model aggregation of the $t$-th communication round, the received signal\cite{WFL} is given by 
\begin{equation}
    \boldsymbol{y}^{\left[t\right]}=\sum_{l=1}^{L}{\boldsymbol{h}_{l} p_l \boldsymbol{s}_l^{\left[t\right]}}+\boldsymbol{n}_0
,\end{equation}
where $p_l\in\mathbb{C}$ is the transmitter scalar of the $l$-th group leader, $\boldsymbol{s}_l\in\mathbb{C}^{1\times q}$ is the gradient aggregated at the $l$-th group leader from its group members, and $\boldsymbol{n}_0\in\mathbb{C}^{N_a\times q}$ is the additive white Gaussian noise (AWGN) with elements following $\mathcal{CN}\left(0,\sigma_{n_0}^2\right)$.

% Assume that the communication interference between clients is not considered.} 
Suppose that all clients follow a CSMA-CA protocol, e.g., the IEEE 802.11 protocol with RTS/CTS, where concurrent transmissions of multiple clients within each other's transmission coverage are prevented in a distributed fashion.
The SNR is used to measure the communication quality, which can vary substantially among the clients due to the geographical distribution of the clients. 
$\gamma_k = p_k |\boldsymbol{h}_{k}|^2 /\sigma_{n_0}^2$ is the received SNR from the $k$-th client to the BS. Likewise, $\gamma_{jk} = p_k |h_{jk}|^2 / \sigma_{n_0}^2$ is the received SNR when the $k$-th client transmits the gradients to the $j$-th client.

\section{Proposed Dual-Segment Clustering Strategy}

\begin{figure}
\centerline{\includegraphics[width=1\linewidth]{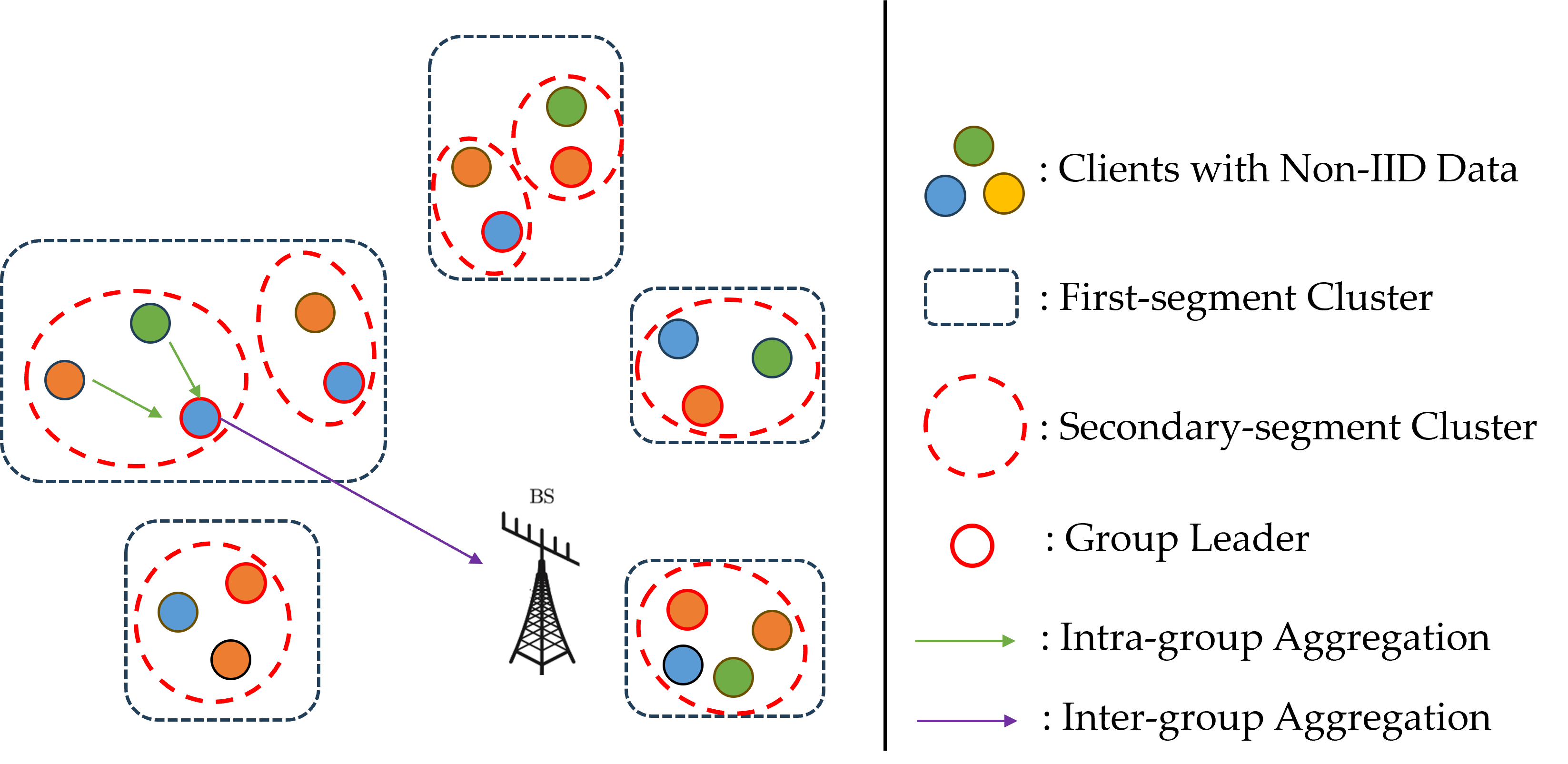}}
\caption{The workflow of the proposed DSC-FL, where each group is expected to contain as many labels as possible, while the communication quality of each client is similar.}
\label{frame}
\end{figure}

In this paper, we develop a new DSC strategy for clients with heterogeneous data and communication conditions. Utilizing the affinity propagation algorithm, we first cluster clients into primary groups based on communication quality, then refine these groups with a novel information quantity matrix to ensure diverse sample labels. The workflow of DSC strategy is illustrated in Fig. \ref{frame}. This strategy effectively mitigates the effects of communication and data heterogeneity on FL convergence, as analyzed in Section III-B.

\subsection{DSC Strategy}

To form primary groups with similar communication quality for accurate local gradient transmission, SNRs are used as the clustering criterion.
Assume that the geographical location and transmission powers of all clients are fixed during the FL process; i.e., their communication quality does not change over rounds. We construct the SNR matrix as
\begin{equation}
    \boldsymbol{\Gamma} = \begin{bmatrix}
        \gamma_1 & \gamma_{12} & \cdots & \gamma_{1K} \\
        \gamma_{21} & \gamma_2 & \cdots & \gamma_{2K} \\
        \vdots & \vdots & \ddots & \vdots \\
        \gamma_{K1} & \gamma_{K2} & \cdots & \gamma_{K} 
    \end{bmatrix},
\label{snr}
\end{equation}
where $\gamma_i$ is the SNR between client $i$ and the BS, while $\gamma_{ij}$ ($i\neq j$) represents the SNR between the $i$-th and the $j$-th clients. $\boldsymbol{\Gamma}$ is a symmetric matrix. 
The affinity propagation algorithm\cite{affinity} determines the number of clusters by identifying exemplars—data points that best represent each cluster.  It begins with a similarity matrix reflecting pairwise similarities and iteratively exchanges responsibility and availability messages, where responsibility indicates a point's suitability as an exemplar and availability reflects the appropriateness of selecting it.  This continues until each point is assigned to the exemplar with the highest combined responsibility and availability, forming the clusters.  The algorithm does not require pre-specified cluster numbers and is well-suited for complex, multi-criteria clustering tasks in heterogeneous environments.  We apply the affinity propagation algorithm to the SNR matrix, efficiently constructing primary clusters with similar communication quality and ensuring unambiguous client grouping.

A similarity matrix $\boldsymbol{S}_{c}$ is constructed to describe the similarity between the clients in communication quality, i.e.,
\begin{equation}
    \boldsymbol{S}_{c} = \begin{bmatrix}
        P_{1} & -\gamma^2_{12} & \cdots & -\gamma^2_{1K} \\
        -\gamma^2_{21} & P_{2} & \cdots & -\gamma^2_{2K} \\
        \vdots & \vdots & \ddots & \vdots \\
        -\gamma^2_{K1} & -\gamma^2_{K2} & \cdots & P_{K} 
    \end{bmatrix},
\label{sc}
\end{equation}
where $\{P_1,\cdots,P_K\}$ collects the preference values for communication quality, implying the likelihood of client $k \in \left\{1,\cdots,\ K\right\}$ being the leader of a group and affecting the number of groups. 

A responsibility matrix $\boldsymbol{R}_{c} (i,k)$ is defined to characterize the likelihood of client $k$ serving as the group leader of client $i$. An attribution matrix $\boldsymbol{A}_{c} (i,k)$ is defined to measure the appropriateness of client $i$ nominating client $k$ as its group leader. Both $\boldsymbol{A}_c$ and $\boldsymbol{R}_c$ are initialized as all-zero matrices.

\begin{algorithm}[t]
\caption{Proposed Dual-Segment Clustering Strategy.}
\begin{algorithmic}[1] % 1表示显示行号，可以根据需要调整
\STATE \textbf{Parameter:}  The  numbers of samples $|\mathcal{D}_k|$ and labels $\mathcal{C}_k^\iota$ of each client, the responsibility matrices $\boldsymbol{R}_{c_0}$ and $\boldsymbol{R}_{d_0}$, the attribution matrices $\boldsymbol{A}_{c_0}$ and $\boldsymbol{A}_{d_0}$, and the learning rate $\lambda$.
% \STATE
\STATE \textbf{Cluster based on the communication quality}:
\STATE \quad Calculate $\boldsymbol{\Gamma}$ by \eqref{snr} and $\boldsymbol{S}_c$ by \eqref{sc}:
\STATE \quad \textbf{for}  $t_{cl} \in T_{cl}$ \textbf{do}
\STATE \quad \quad Calculate $\boldsymbol{R}_c$ by \eqref{rc}, $\boldsymbol{A}_c$ by \eqref{ac1} and \eqref{ac2}.
\STATE \quad \textbf{end for}
\STATE \quad \textbf{Return} $l_{com}$ primary groups.
\STATE \quad \textbf{Data-based Cluster within $l_{com}$ primary groups}: 
\STATE \quad \quad Calculate $\boldsymbol{\Xi}$ by \eqref{datama} and $\boldsymbol{S}_d$ by \eqref{sd}:
\STATE \quad \quad \textbf{for} $t_{cl} \in T_{cl}$ \textbf{do}
\STATE \quad \quad \quad Calculate $\boldsymbol{R}_d$ like \eqref{rc}, $\boldsymbol{A}_d$ like \eqref{ac1} and \eqref{ac2}.
\STATE \quad \quad \textbf{end for}
% \STATE
\FOR{$t \leftarrow 0,1,2,\dots , T$}
    \STATE Aggregation by \eqref{aggregationw}.
    \STATE Update the global model by \eqref{global} and broadcast the global model to the clients.
\ENDFOR
\STATE \textbf{Return} $\boldsymbol{w}$.
\end{algorithmic}
\end{algorithm}

This clustering algorithm iterates over $\boldsymbol{R}_{c} (i,k)$ and $\boldsymbol{A}_{c} (i,k)$ based on the affinity propagation algorithm until the group boundaries do not change for $T_{cl}$ consecutive rounds. Particularly, the responsibility information is updated by 
\begin{equation}
    \boldsymbol{R}_{c}(i,k) = \boldsymbol{S}_{c}(i,k)-\max_{k \neq k'}{[\boldsymbol{S}_{c}(i,k')+\boldsymbol{A}_{c}(i,k')]}.
\label{rc}
\end{equation}
The attribution information is updated by
\begin{equation}
    \boldsymbol{A}_{c}(i,k) \!= \!\min[0,\boldsymbol{R}_{c}(k,k)\!\!+\!\!\!\!\sum_{i' \notin (i,k)}{\max(0,\boldsymbol{R}_{c}(i',k))}],i\!\!\neq \!\! k,
\label{ac1}
\end{equation}
and 
\begin{equation}
    \boldsymbol{A}_{c}(i,i) = \max_{i' \neq k}{[0,\boldsymbol{R}_{c}(i',k)]},i=k.
\label{ac2}
\end{equation}

% Finally, the responsibility information and attribution information are used to jointly determine the group leaders and members. Specifically, if the maximum element of the $i$-th row in $\boldsymbol{C}_{c}=\boldsymbol{R}_{c}(i,k)+\boldsymbol{A}_{c}(i,k)$ is in the diagonal position, the $i$-th client is the group leader corresponding to the column index. Otherwise, the $i$-th client is a group member indexed by the column corresponding to the maximum element.

The responsibility information and attribution information jointly determine the group leaders and members. Specifically, for the $i$-th client, we examine the $i$-th row of the combined matrix $\boldsymbol{C}_{c}=\boldsymbol{R}_{c}(i,k)+\boldsymbol{A}_{c}(i,k)$.  If the maximum of this row is located on the diagonal, the $i$-th client is designated as the group leader corresponding to the column index. If the maximum is not on the diagonal, the $i$-th client is classified as a group member, with the corresponding group leader identified by the column index of the maximum element.

After clustering based on the communication quality, the communication conditions are reasonably consistent within the group. Next, the clients are further clustered according to data heterogeneity within each primary group, so that the clients can contain as many classes of sample labels as possible in each secondary cluster. 
% This is achieved by constructing an information matrix for the clients' data.

Suppose that $K$ clients in the FL system possess a total of $\mathcal{D}$ data samples and $\mathcal{L}$ labels. The number of data samples with the $\iota$-th label of the $k$-th client is $\mathcal{C}_k^\iota$. The total number of samples with the $\iota$-th label is $\mathcal{C}^\iota$. The probability that a sample belongs to the $\iota$-th class label in the dataset of the $k$-th client is $P_1 =\mathcal{C}_k^\iota/\mathcal{D}$. The probability of its belonging to the $k$-th client is $P_2 =\mathcal{D}_k/\mathcal{D}$. The probability of its belonging to the $\iota$-th class label is $P_3 =\mathcal{C}^\iota/\mathcal{D}$. Then, a matrix measuring the distribution of the dataset can be written as
\begin{equation}
    \boldsymbol{\Xi} = -\begin{bmatrix}
        \frac{\mathcal{C}_1^1}{\mathcal{D}}\log{\frac{\mathcal{D}\mathcal{C}_1^1}{\mathcal{D}_1\mathcal{C}^1}} & \frac{\mathcal{C}_1^2}{\mathcal{D}}\log{\frac{\mathcal{D}\mathcal{C}_1^2}{\mathcal{D}_1\mathcal{C}^2}} & \cdots & \frac{\mathcal{C}_1^\mathcal{L}}{\mathcal{D}}\log{\frac{\mathcal{D}\mathcal{C}_1^\mathcal{L}}{\mathcal{D}_1\mathcal{C}^\mathcal{L}}} \\
        \frac{\mathcal{C}_2^1}{\mathcal{D}}\log{\frac{\mathcal{D}\mathcal{C}_2^1}{\mathcal{D}_2\mathcal{C}^1}} & \frac{\mathcal{C}_2^2}{\mathcal{D}}\log{\frac{\mathcal{D}\mathcal{C}_2^2}{\mathcal{D}_2\mathcal{C}^2}} & \cdots & \frac{\mathcal{C}_2^\mathcal{L}}{\mathcal{D}}\log{\frac{\mathcal{D}\mathcal{C}_2^\mathcal{L}}{\mathcal{D}_2\mathcal{C}^\mathcal{L}}} \\
        \vdots & \vdots & \ddots & \vdots \\
        \frac{\mathcal{C}_K^1}{\mathcal{D}}\log{\frac{\mathcal{D}\mathcal{C}_K^1}{\mathcal{D}_K\mathcal{C}^1}} & \frac{\mathcal{C}_K^2}{\mathcal{D}}\log{\frac{\mathcal{D}\mathcal{C}_K^2}{\mathcal{D}_K\mathcal{C}^2}} & \cdots & \frac{\mathcal{C}_K^\mathcal{L}}{\mathcal{D}}\log{\frac{\mathcal{D}\mathcal{C}_K^\mathcal{L}}{\mathcal{D}_K\mathcal{C}^\mathcal{L}}} 
    \end{bmatrix},
\label{datama}
\end{equation}
where $\frac{\mathcal{C}_k^\iota}{\mathcal{D}}\log{\frac{\mathcal{D}\mathcal{C}_k^\iota}{\mathcal{D}_k\mathcal{C}^\iota}}$ quantifies the information that a sample is classified into the $\iota$-th label of the $k$-th client. Based on $\boldsymbol{\Xi}$, we construct a similarity matrix to describe the distribution of the dataset, as given by
\begin{equation}
    \boldsymbol{S}_{d} = \begin{bmatrix}
        P_{d} & s_{1,2} & \cdots & s_{1,K_{l_{com}}} \\
        s_{2,1} & P_{d} & \cdots & s_{2,K_{l_{com}}} \\
        \vdots & \vdots & \ddots & \vdots \\
        s_{K_{l_{com}},1} & s_{K_{l_{com}},2} & \cdots & P_{d} 
    \end{bmatrix},
\label{sd}
\end{equation}
where $s_{i,k}=\{\sum_{\iota=1}^{\mathcal{L}}{[\boldsymbol{\Xi} (i,\iota)-\boldsymbol{\Xi} (k,\iota)]^2}\}^2$ with $i \neq k$, ${P}_{d}$ is the preference value for the data distribution, and $K_{l_{com}}$ denotes the number of clients in the $l_{com}$-th primary group.

Similarly, we define a responsibility matrix $\boldsymbol{R}_{d} (i,k)$ and an attribution matrix $\boldsymbol{A}_{d} (i,k)$ for the similarity matrix $\boldsymbol{S}_d$ to iteratively update the secondary groups based on the data heterogeneity. The group leaders and members are selected in the same way as in the primary groups. For the remaining ungrouped clients, the Euclidean distance-based proximity principle can be adopted to group them into their respective nearby groups. 
\textbf{Algorithm 1} describes the proposed DSC strategy. 

\subsection{Convergence analysis}

% This subsection analyzes the convergence of the proposed DSC strategy. 
The effectiveness of the DSC strategy is assessed by analyzing the impact of data and communication heterogeneity on FL convergence.
Four assumptions are made to facilitate the convergence analysis\cite{cross}:

\textbf{A1.} $\nabla F\left(\boldsymbol{w}\right)$ satisfies uniformly $L$-Lipschitz continuous with regard to the model parameter $\boldsymbol{w}$, i.e., $\|\nabla F\left(\boldsymbol{w}^{\left[t+1\right]}\right) - \nabla F\left(\boldsymbol{w}^{\left[t\right]}\right)\| \leq L\|\boldsymbol{w}^{\left[t+1\right]}-\boldsymbol{w}^{\left[t\right]}\|$. 

\textbf{A2.} $F\left(\boldsymbol{w}\right)$ is a strongly convex function of $\boldsymbol{w}$ with the parameter $\mu>0$, i.e., $\mathcal{F}\left(\boldsymbol{w}^{\left[n+1\right]}\right) \geq \mathcal{F}\left(\boldsymbol{w}^{\left[n\right]}\right)
    +\left(\boldsymbol{w}^{\left[n+1\right]}-\boldsymbol{w}^{\left[n\right]}\right)^T\cdot
    \nabla \mathcal{F}\left(\boldsymbol{w}^{\left[n\right]}\right)+\frac{\mu}{2}\|\boldsymbol{w}^{\left[n+1\right]}-\boldsymbol{w}^{\left[n\right]}\|^2$. 

\textbf{A3.} $F\left(\boldsymbol{w}\right)$ is second-order continuously differentiable. 

\textbf{A4.} The local loss function $F_k\left(\boldsymbol{w}^{\left[t\right]}\right)$ is $\delta$-locally dissimilar at $\boldsymbol{w}^{[t]}$, i.e., $\mathbb{E}[\|\nabla F_k(\boldsymbol{w}^{[t]})\|^2] \leq \|\nabla F(\boldsymbol{w}^{[t]})\|^2 \delta ^2$, where the dissimilarity factor $\delta \geq 1$ describes the heterogeneity degree of the data distribution. 
The ensuing theorem delineates the convergence of FL under the DSC strategy.

\vspace{2 mm}
\noindent\textbf{Theorem 1: }\textit{Given the optimal global model $\boldsymbol{w}^\ast$ under the ideal channel condition,  the intra-group dissimilarity factors $\delta_{intra}$, the inter-group dissimilarity factors $\delta_{inter}$, the intra-group communication impact factor $\sigma_k$, the inter-group communication impact factor $\sigma_l$, and the learning rate $\lambda$,  the convergence upper bound of FL is given 
% in \eqref{theorem}, 
by
% \begin{figure*}[t]
    \begin{equation}
    \begin{aligned}
\mathbb{E}&\left[F\left(\boldsymbol{w}^{\left[t+1\right]}\right)-F\left(\boldsymbol{w}^\ast\right)\right]\le A^T\mathbb{E}\left[F\left(\boldsymbol{w}^{\left[0\right]}\right)-F\left(\boldsymbol{w}^\ast\right)\right]\\
&\quad +\frac{L \lambda^2}{2}(\sum_{l=1}^L{G_l^2 \sigma_l^2}+\sum_{l=1}^L{\sum_{k=1}^{K_l}{G_k^2 \sigma_k^2}})\cdot\frac{1-A^N}{1-A},
\end{aligned}
    \label{theorem}
    \end{equation}
% \end{figure*}
where $A=1+\mu L \lambda^2 \delta_{inter}^2 \sum_{l=1}^L{G_l^2 (\sum_{k=1}^{K_l}{G_k^2})\delta_{intra}^2}-2\mu \lambda$, and $T$ is the total number of aggregations.} 

\textit{Proof}: See Appendix~I in the supplementary file.
\vspace{2 mm}

By \textbf{Theorem 1},  the upper bound of $\mathbb{E}\left[F\left(\boldsymbol{w}^{\left[t+1\right]}\right)-F\left(\boldsymbol{w}^\ast\right)\right]$ converges at the rate $A<1$; i.e., FL surely converges when the learning rate $\lambda$ satisfies
\begin{equation}
    \lambda < \frac{2}{L \delta_{inter}^2 \sum_{l=1}^L{G_l^2 (\sum_{k=1}^{K_l}{G_k^2})\delta_{intra}^2}} .
\end{equation}
The learning rate $\lambda$ needs to be inversely proportional to the heterogeneity degree of the data (measured by $\delta_{inter}^2$ and $\delta_{intra}^2$). The more heterogeneous the data distribution, the smaller $\lambda$ is needed to ensure convergence, resulting in slower convergence.

% Under the proposed DSC strategy, the use of the affinity propagation algorithm upon \eqref{sd} makes the data distribution between the secondary groups close to each other, i.e., $\delta_{inter}^2$ is close to 1, so that a larger learning rate $\lambda$ can be obtained and the convergence rate can be accelerated. Meanwhile, the gradient direction of the intra-group aggregation is more consistent due to $K_l<K$, although a certain degree of Non-IID aggregation is allowed within each group. Thus, $\lambda$ can be set to a larger value, thus increasing the convergence rate. 

% As for the upper bound of \eqref{theorem}, the clients' SNRs in each group are reasonably consistent, making the communication error term small, i.e. $\sigma_k^2\approx0$, and hence reducing the error of gradient update. Although the SNRs between the group leaders can be different, the reasonably consistent inter-groups data distributions are conductive to the consistency of the gradients. The error term $\sigma_l^2$ would be diluted by the weighted aggregation and would not significantly affect the overall convergence direction.
The proposed DSC strategy reduces the errors in gradient update and communication by balancing data heterogeneity and transmission capability, enabling wireless FL to achieve faster and more stable convergence.
On the one hand, the application of the affinity propagation algorithm to \eqref{sd} effectively reduces the inter-group data distribution disparity, leading to $\delta_{inter}^2$ approaching 1. This directly influences the selection of $\lambda$ by allowing a larger $\lambda$ to be selected, which consequently accelerates the convergence. Furthermore, $K_l < K$ ensures greater consistency in the gradient directions during the intra-group aggregation, even in the presence of a certain degree of Non-IID aggregation within each group. This consistency further enhances convergence.

On the other hand, clustering upon \eqref{snr}  ensures that the SNRs of the clients within each group are relatively consistent, thereby minimizing the communication error, i.e., $\sigma_k^2 \approx 0$ in \eqref{theorem}, which reduces the error in the gradient updates. Although there may be variations in SNRs among the group leaders, the relatively consistent inter-group data distributions contribute to maintaining more similar gradient. The error $\sigma_l^2$ in \eqref{theorem} can be effectively mitigated by the weighted aggregation and would not significantly affect the overall convergence direction.

% In summary, the proposed DSC strategy reduces the error in the gradient update and communication by balancing data heterogeneity and transmission capability, thus speeding up the global convergence, especially under the Non-IID condition. Wireless FL systems under the DSC strategy not only converge faster, but also maintain a stable convergence process even when the transmission conditions are substantially different.

% The algorithm iteratively updates the responsibility matrix $\boldsymbol{R}$ and the attribution matrix $\boldsymbol{A}$ based on the communication quality and data distribution, respectively, in an alternating manner until the specified iteration round $T_{cl}$ is reached.
% The SNRs and data distribution of the $K$ clients are used to initialize these two matrices with the computational complexity of $\mathcal{O}(K^2)$. 
% In each iteration of the algorithm, the complexity is dominated by the evaluation of $\boldsymbol{R}$ and $\boldsymbol{A}$. Thus, the overall complexity of the DSC algorithm is $\mathcal{O}(T_{cl}\cdot K^2)$, or $\mathcal{O}(K^2)$ since $T_{cl}$ is constant.

\section{Simulation Results}

\subsection{Simulation Setup and Baselines}

Consider a rectangular area with a side length of 100 meters. 50 clients and a BS serving as the parameter server are randomly distributed in the area. With reference to~\cite{yuanxiaojun}, the path loss model is $PL_{DB}=G_{BS}G_D {(\frac{c}{4\pi f_c d_{DB}})}^P$, where the antenna gain is $G_{BS}=5$ dBi at the BS and $G_D=0$ dBi at clients, $f_c=915$ MHz is the carrier frequency, $P=3.76$ is path loss exponent, $d$ is the distance, and $c$ is the speed of light. The transmit power of the clients is $0.1$ W. The noise power is $0.001$ W. We use a CNN network with two $5\times 5$ convolution layers (each with $2\times2$ max pooling), followed by a batch normalization layer, a fully connected layer with 50 units, a ReLu activation layer, and a softmax output layer. We train and test on the MNIST and Fashion-MNIST datasets. The data samples are randomly distributed among the clients, each assigned 400 to 800 samples of two random labels. The SGD algorithm with $batchsize=0.1$ is used to train the local models. The learning rate is $\lambda=0.06$ for the MNIST dataset and $\lambda=0.05$ for the Fashion-MNIST dataset.

We test the DSC strategy (\textbf{Setting 1}), the data-based clustering part of the DSC strategy (\textbf{Setting 2}) and the communication-based clustering part of the DSC strategy (\textbf{Setting 3}). For comparison, we set two baselines: 1) The state-of-the-art clustering algorithm (\textbf{Benchmark 1}) developed in \cite{GFedAvg}, named GFedAvg, where the sparsity of the clients' labels and their Euclidean distances are exploited, without the consideration of communication quality, and 2) The most widely used FedAvg algorithm (\textbf{Benchmark 2})\cite{FedAvg}, where each client directly uploads model parameters without grouping. Its data size serves as its aggregation weight.

\subsection{Effectiveness of DSC-FL}

The five considered schemes are tested in an ideal communication environment (without noise) and a noisy communication environment. \textbf{Setting 2} is not tested in the ideal communication environment due to the fact that there is no need for clustering clients based on their communication qualities in that environment.

\begin{figure}
    \begin{minipage}{0.4995\linewidth}
        \centering
        \includegraphics[width=1.0\linewidth]{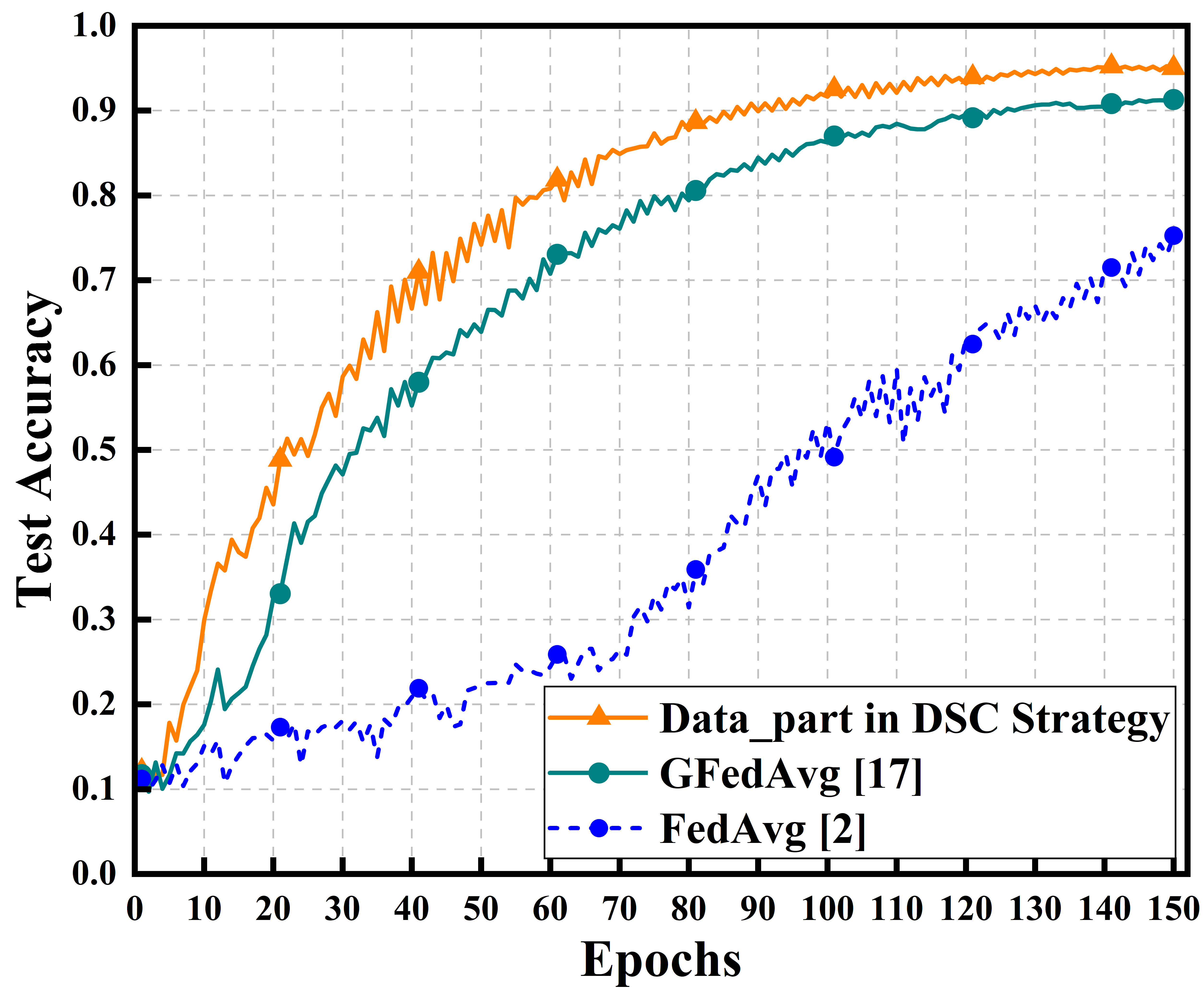}
        \subcaption{MNIST}
    \end{minipage}%
    \begin{minipage}{0.4995\linewidth}
        \centering
        \includegraphics[width=1.0\linewidth]{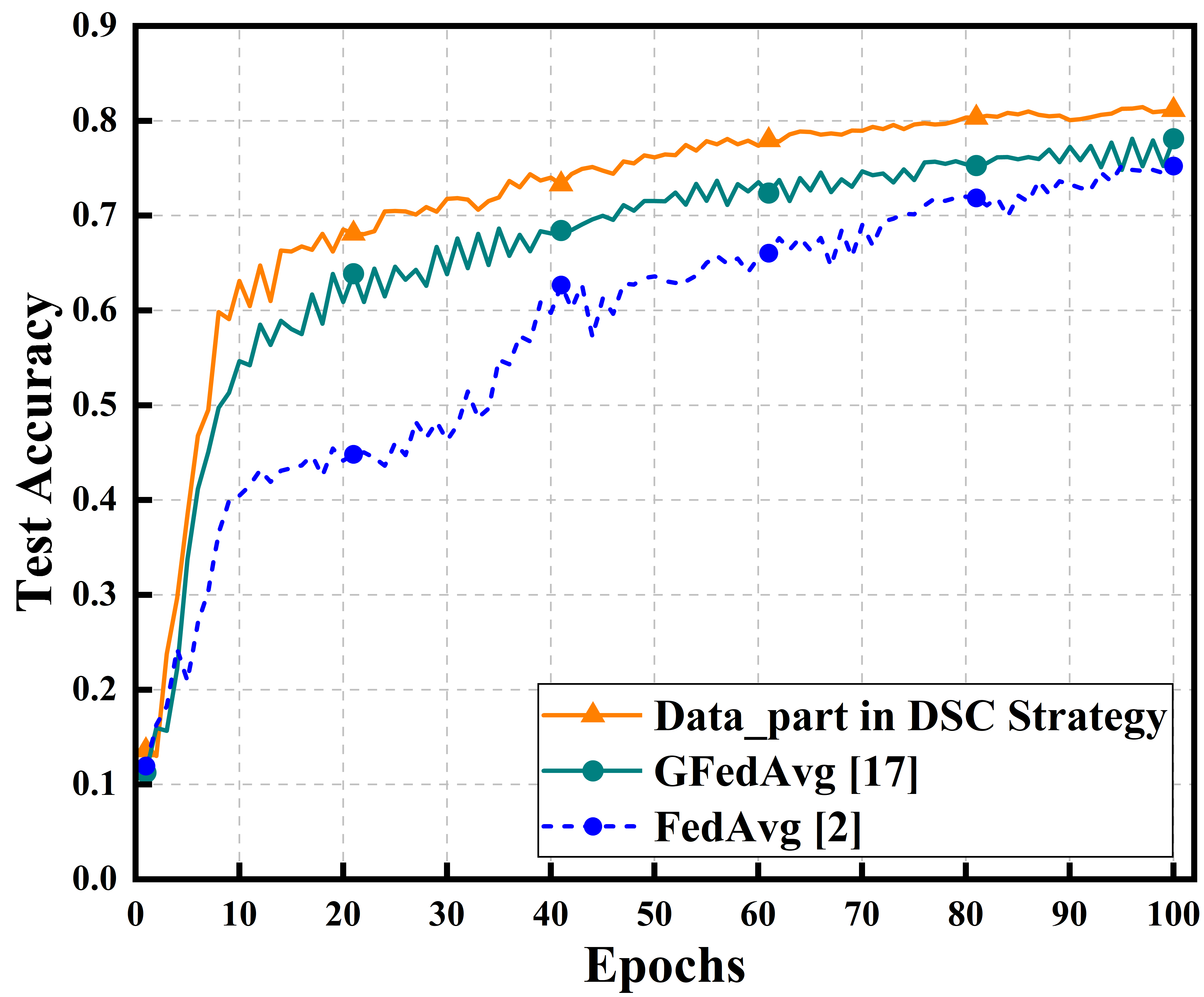}
        \subcaption{Fashion-MNIST}
    \end{minipage}
    \caption{The performance of FL after clustering in an ideal communication environment.}
    \label{ideal}
\end{figure}

% \begin{figure}
%     \begin{subfigure}{\columnwidth} % 使用整个一列的宽度
%         \centering
%         \includegraphics[width=0.9\columnwidth]{MNIST_Airfree.png} % 子图宽度占整列宽度的一部分
%         \caption{MNIST}
%         \includegraphics[width=0.9\columnwidth]{FMNIST_Airfree.png} % 子图宽度占整列宽度的一部分
%         \caption{Fashion-MNIST}
%     \end{subfigure}
%     \caption{The performance of FL after clustering in an ideal communication environment.}
% \label{ideal}
% \end{figure}

Fig. \ref{ideal} shows the performance of \textbf{Setting 2} and \textbf{Benchmarks 1 and 2} in the ideal communication environment. 
\textbf{Setting 2} outperforms FedAvg algorithm \textbf{(Benchmark 2)} on both two datasets by 19.74\% (MNIST) or 5.91\% (Fashion-MNIST), demonstrating the effectiveness of the data-based clustering part of the proposed DSC strategy. 
Compared with the existing algorithm (\textbf{Benchmark 1}), the proposed algorithm is better by 3.71\% (MNIST) or 3.03\% (Fashion-MNIST), in testing accuracy in the ideal communication environment.
% , demonstrating the superiority of the data-based clustering part of the proposed DSC strategy. 

\begin{figure}
    \begin{minipage}{0.4995\linewidth}
        \centering
        \includegraphics[width=1.0\linewidth]{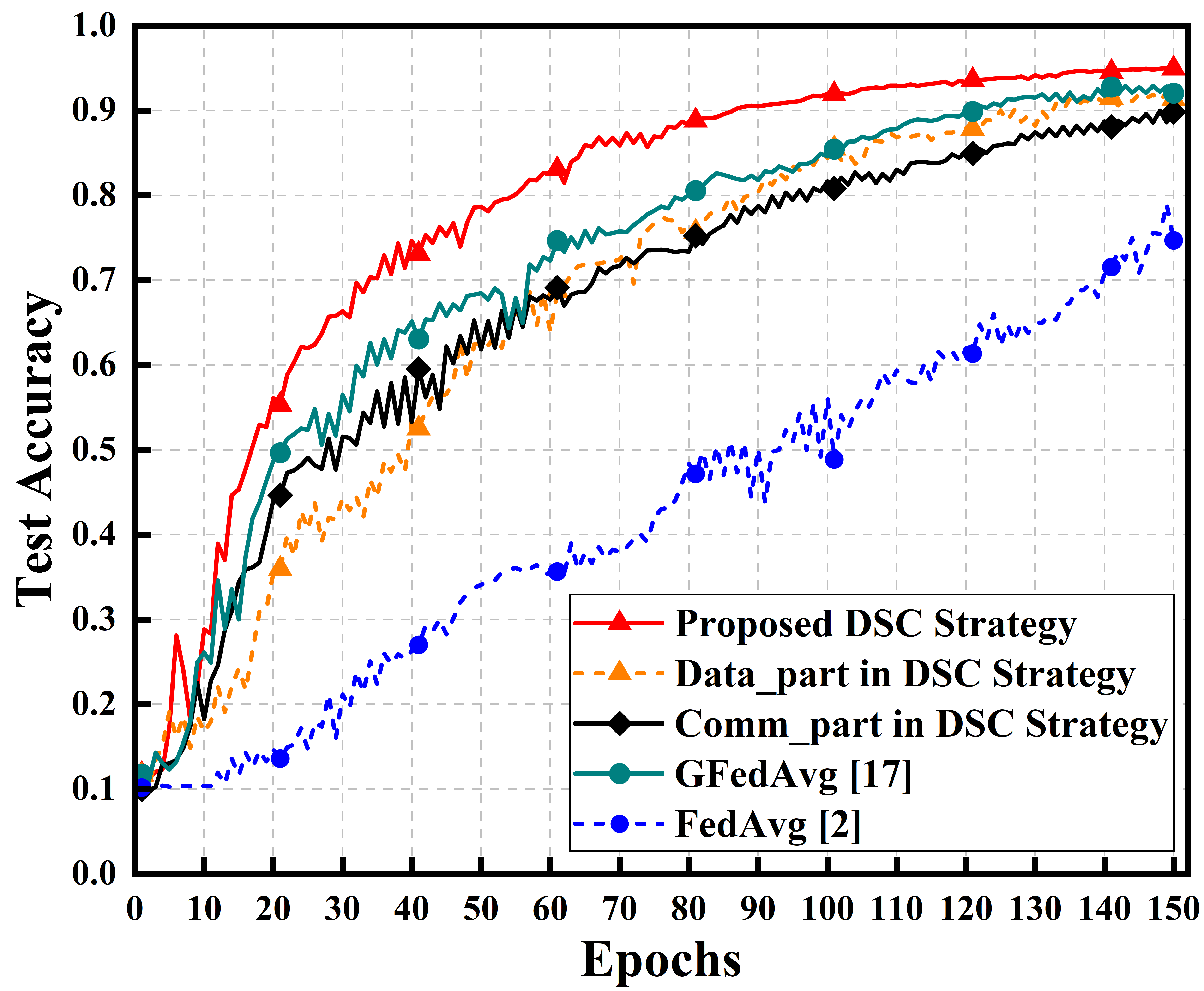}
        \subcaption{MNIST}
    \end{minipage}%
    \begin{minipage}{0.4995\linewidth}
        \centering
        \includegraphics[width=1.0\linewidth]{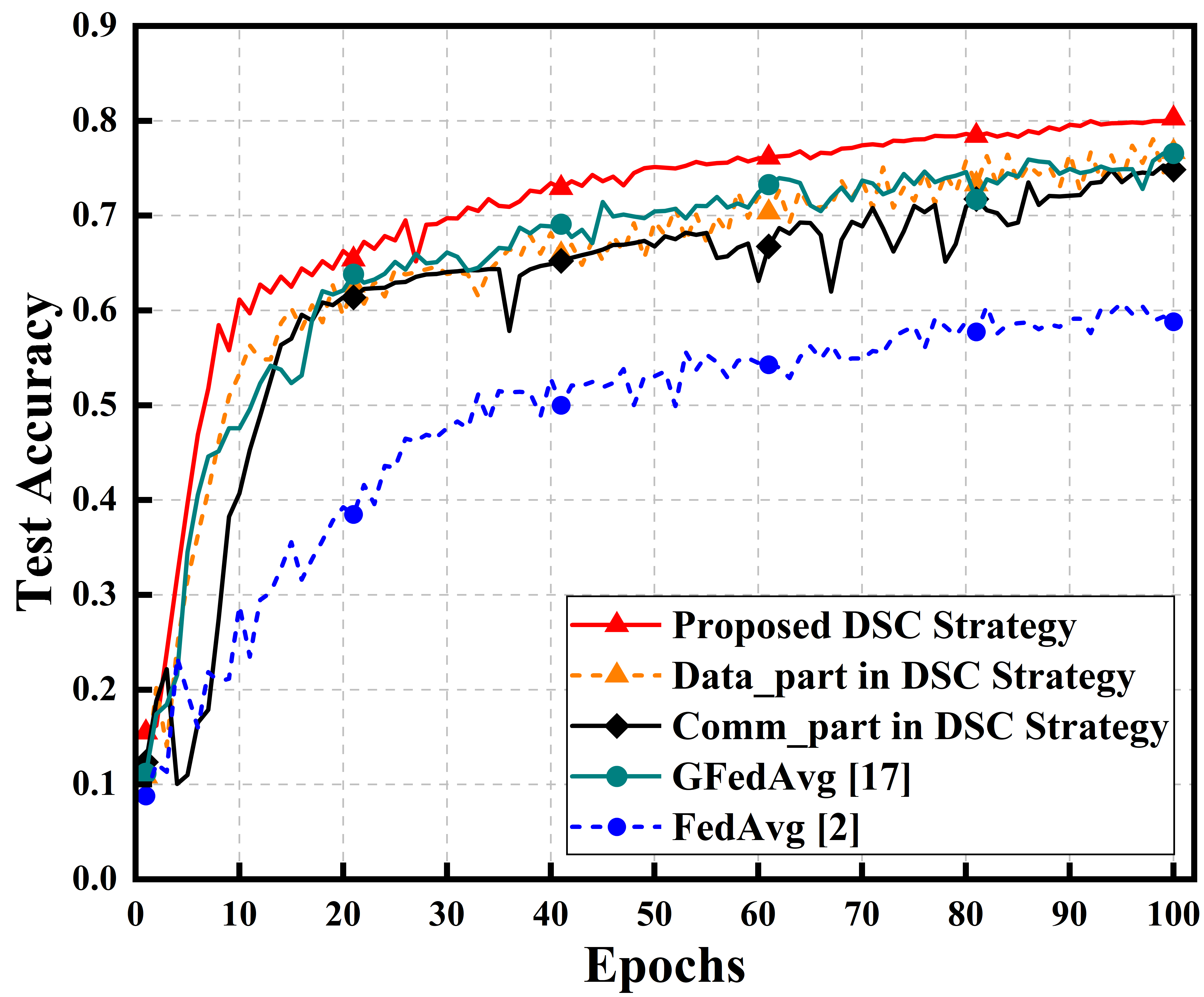}
        \subcaption{Fashion-MNIST}
    \end{minipage}
    \caption{The performance of FL after clustering in a practical communication environment.}
    \label{actual}
\end{figure}

% \begin{figure}
%     \begin{subfigure}{\columnwidth} % 使用整个一列的宽度
%         \centering
%         \includegraphics[width=0.9\columnwidth]{MNIST_Noise.png} % 子图宽度占整列宽度的一部分
%         \caption{MNIST}
%         \includegraphics[width=0.9\columnwidth]{FMNIST_Noise.png} % 子图宽度占整列宽度的一部分
%         \caption{Fashion-MNIST}
%     \end{subfigure}
%     \caption{The performance of FL after clustering in a practical communication environment.}
% \label{actual}
% \end{figure}

Fig. \ref{actual} plots the testing accuracy of the five considered schemes in the noisy communication environment. 
% Compared with FedAvg algorithm \textbf{(Benchmark 2)}, the DSC strategy (\textbf{Setting 1}) and its data-based clustering part (\textbf{Setting 2}) are substantially better by 20.28\% (MNIST) and 21.42\% (Fashion-MNIST), and 16.66\% (MNIST) and 17.89\% (Fashion-MNIST), respectively. 
Compared with FedAvg algorithm \textbf{(Benchmark 2)}, the DSC strategy (\textbf{Setting 1}) is substantially better by 20.28\% (MNIST) or 21.42\% (Fashion-MNIST). The data-based clustering part (\textbf{Setting 2}) is better by 16.66\% (MNIST) and 17.89\% (Fashion-MNIST). The effectiveness of the proposed DSC strategy is confirmed in the noisy communication environment.
Moreover, the full DSC strategy \textbf{(Setting 1)} improves the testing accuracy by 2.92\% on MNIST and 3.68\% on Fashion-MNIST, compared with the GFedAvg \textbf{(Benchmark 1)}, although its data-based clustering part \textbf{(Setting 2)} is not much better than the GFedAvg in noisy environments. 
This is because the GFedAvg may cluster the same clients into multiple groups at the same time, resulting in repeated uploading of model parameters, which compensates for FL performance to some extent. The data-based clustering step of the proposed DSC strategy can cause performance degradation in a noisy communication environment, but the full DSC strategy can overcome this. Moreover, the communication-based clustering part (\textbf{Setting 3}) performs worse than the GFedAvg \textbf{(Benchmark 1)} and the data-based clustering part \textbf{(Setting 2)}, indicating that the impact of communication heterogeneity on wireless FL is weaker than that of data distribution heterogeneity, but cannot be ignored. Therefore, the proposed DSC strategy, which comprehensively considers the effects of both data and communication heterogeneity on wireless FL, is critical.
The importance and superiority of the DSC strategy are demonstrated.

\section{Conclusion}

In this paper, a new DSC strategy was proposed to address data and communication heterogeneity in wireless FL. Extensive simulations indicate that the strategy can improve testing accuracy by 20.28\% on MNIST, and by 21.42\% on Fashion-MNIST in a heterogeneous network condition. Our future work will focus on optimal clustering and resource configurations in time-varying mobile environments.

\bibliographystyle{IEEEtran}
\bibliography{references}\ %IEEEabrv instead of IEEEfull

\begin{thebibliography}{10}
\providecommand{\url}[1]{#1}
\csname url@samestyle\endcsname
\providecommand{\newblock}{\relax}
\providecommand{\bibinfo}[2]{#2}
\providecommand{\BIBentrySTDinterwordspacing}{\spaceskip=0pt\relax}
\providecommand{\BIBentryALTinterwordstretchfactor}{4}
\providecommand{\BIBentryALTinterwordspacing}{\spaceskip=\fontdimen2\font plus
\BIBentryALTinterwordstretchfactor\fontdimen3\font minus \fontdimen4\font\relax}
\providecommand{\BIBforeignlanguage}[2]{{%
\expandafter\ifx\csname l@#1\endcsname\relax
\typeout{** WARNING: IEEEtran.bst: No hyphenation pattern has been}%
\typeout{** loaded for the language `#1'. Using the pattern for}%
\typeout{** the default language instead.}%
\else
\language=\csname l@#1\endcsname
\fi
#2}}
\providecommand{\BIBdecl}{\relax}
\BIBdecl

\bibitem{FL1}
C.~Huang, E.~Liu, R.~Wang, Y.~Liu, H.~Zhang, Y.~Geng, J.~Wang, and S.~Han, ``Personalized federated learning via directed acyclic graph based blockchain,'' \emph{IET Blockchain}, vol.~4, no.~1, pp. 73--82, 2024.

\bibitem{FedAvg}
B.~McMahan, E.~Moore, D.~Ramage, S.~Hampson, and B.~A. y~Arcas, ``Communication-efficient learning of deep networks from decentralized data,'' in \emph{AISTATS}, pp. 1273--1282.\hskip 1em plus 0.5em minus 0.4em\relax PMLR, 2017.

\bibitem{noniid2}
K.~Hsieh, A.~Phanishayee, O.~Mutlu, and P.~Gibbons, ``The non-iid data quagmire of decentralized machine learning,'' in \emph{ICML}, pp. 4387--4398.\hskip 1em plus 0.5em minus 0.4em\relax PMLR, 2020.

\bibitem{noniidcomm}
M.~Shirvanimoghaddam, A.~Salari, Y.~Gao, and A.~Guha, ``Federated learning with erroneous communication links,'' \emph{IEEE Commun. Lett.}, vol.~26, no.~6, pp. 1293--1297, 2022.

\bibitem{hierarchical}
M.~Duan, D.~Liu, X.~Chen, R.~Liu, Y.~Tan, and L.~Liang, ``Self-balancing federated learning with global imbalanced data in mobile systems,'' \emph{IEEE Trans. Parallel Distrib. Syst.}, vol.~32, no.~1, pp. 59--71, 2020.

\bibitem{dataquantization}
S.~Seo, J.~Lee, H.~Ko, and S.~Pack, ``Performance-aware client and quantization level selection algorithm for fast federated learning,'' in \emph{IEEE WCNC}, pp. 1892--1897.\hskip 1em plus 0.5em minus 0.4em\relax IEEE, 2022.

\bibitem{comgroup1}
L.~Liu, J.~Zhang, S.~Song, and K.~B. Letaief, ``Client-edge-cloud hierarchical federated learning,'' in \emph{IEEE ICC}, pp. 1--6.\hskip 1em plus 0.5em minus 0.4em\relax IEEE, 2020.

\bibitem{comgroup2}
C.~Wang, Y.~Yang, and P.~Zhou, ``Towards efficient scheduling of federated mobile devices under computational and statistical heterogeneity,'' \emph{IEEE Trans. Parallel Distrib. Syst.}, vol.~32, no.~2, pp. 394--410, 2020.

\bibitem{comdata1}
J.-w. Lee, J.~Oh, Y.~Shin, J.-G. Lee, and S.-Y. Yoon, ``Accurate and fast federated learning via iid and communication-aware grouping,'' \emph{arXiv preprint arXiv:2012.04857}, 2020.

\bibitem{comdata2}
Y.~Lei, L.~Yanyan, C.~Jiannong, H.~Jiaming, and Z.~Mingjin, ``E-tree learning: A novel decentralized model learning framework for edge ai,'' \emph{IEEE Internet Things J.}, vol.~8, no.~14, pp. 11\,290--11\,304, 2021.

\bibitem{comdata3}
Z.~He, L.~Yang, W.~Lin, and W.~Wu, ``Improving accuracy and convergence in group-based federated learning on non-iid data,'' \emph{IEEE Trans. Netw. Sci. Eng.}, vol.~10, no.~3, pp. 1389--1404, 2022.

\bibitem{affinity}
B.~J. Frey and D.~Dueck, ``Clustering by passing messages between data points,'' \emph{Science}, vol. 315, no. 5814, pp. 972--976, 2007.

\bibitem{aggragatewei}
H.~Wu and P.~Wang, ``Fast-convergent federated learning with adaptive weighting,'' \emph{IEEE Trans. Cogn. Commun. Netw.}, vol.~7, no.~4, pp. 1078--1088, 2021.

\bibitem{WFL}
M.~Chen, Z.~Yang, W.~Saad, C.~Yin, H.~V. Poor, and S.~Cui, ``A joint learning and communications framework for federated learning over wireless networks,'' \emph{IEEE Trans. Wirel. Commun.}, vol.~20, no.~1, pp. 269--283, 2020.

\bibitem{cross}
P.~Sun, E.~Liu, W.~Ni, R.~Wang, Z.~Xing, B.~Li, and A.~Jamalipour, ``Reconfigurable intelligent surface-assisted wireless federated learning with imperfect aggregation,'' \emph{IEEE Trans. Commun.}, pp. 1--14, 2024, {Early Access, DOI: 10.1109/TCOMM.2024.3450605}.

\bibitem{yuanxiaojun}
H.~Liu, X.~Yuan, and Y.-J.~A. Zhang, ``Reconfigurable intelligent surface enabled federated learning: A unified communication-learning design approach,'' \emph{IEEE Trans. Wirel. Commun.}, vol.~20, no.~11, pp. 7595--7609, 2021.

\bibitem{GFedAvg}
W.~Nie, L.~Yu, and Z.~Jia, ``Research on aggregation strategy of federated learning parameters under non-independent and identically distributed conditions,'' in \emph{ICAML 2022}, pp. 41--48.\hskip 1em plus 0.5em minus 0.4em\relax IEEE, 2022.

\end{thebibliography}

\vfill

\end{document}